%% file: naacl2021.tex
\pdfoutput=1

\documentclass[11pt]{article}

\usepackage[]{naacl2021}

\usepackage{times}
\usepackage{latexsym}

\usepackage[T1]{fontenc}

\usepackage[utf8]{inputenc}
\usepackage{footnote}
\usepackage{times}
\usepackage{latexsym}

\usepackage{microtype}
\usepackage{multirow}
\usepackage{amsmath} 
\usepackage{booktabs}
\usepackage{graphicx}
\usepackage{import}
\usepackage{comment}
\usepackage{tabularx}
\usepackage{enumitem}
\usepackage{lipsum}  
\usepackage{multicol}

\title{Controllable Text Simplification with Explicit Paraphrasing}

\author{Mounica Maddela\textsuperscript{1}, Fernando Alva-Manchego\textsuperscript{2}, Wei Xu\textsuperscript{1} \\  
\textsuperscript{1}School of Interactive Computing, Georgia Institute of Technology\\
  \textsuperscript{2}Department of Computer Science, University of Sheffield \\
  {\tt \small \{mounica.maddela, wei.xu\}@cc.gatech.edu \quad f.alva@sheffield.ac.uk}\\
}

\begin{document}
\maketitle
\begin{abstract}
Text Simplification improves the readability of sentences through several rewriting transformations, such as lexical paraphrasing, deletion, and splitting. Current simplification systems are predominantly sequence-to-sequence models that are trained end-to-end to perform all these operations simultaneously. However, such systems limit themselves to mostly deleting words and cannot easily adapt to the requirements of different target audiences. In this paper, we propose a novel hybrid approach that leverages linguistically-motivated rules for splitting and deletion, and couples them with a neural paraphrasing model to produce varied rewriting styles.  We introduce a new data augmentation method to improve the paraphrasing capability of our model. Through automatic and manual evaluations, we show that our proposed model establishes a new state-of-the art for the task, paraphrasing more often than the existing systems, and can control the degree of each simplification operation applied to the input texts.\footnote{ Our code and data are available at \url{https://github.com/mounicam/controllable_simplification}.}
\end{abstract}
 
\section{Introduction}
\import{}{01_introduction.tex}
\section{Our Approach}
\label{sec:method}
\import{}{03_model.tex}

\section{Experiments}
\import{}{04_evaluation.tex}

\section{Related Work}
\import{}{06_related_work.tex}

\section{Conclusion}
\import{}{07_conclusion.tex}

\section*{Acknowledgments}

We thank the anonymous reviewers for their valuable  feedback. We thank Newsela for sharing the data and NVIDIA for providing GPU computing resources. This research is supported in part by the NSF award IIS-1822754, ODNI and IARPA via the BETTER program contract 19051600004. The views and conclusions contained herein are those of the authors and should not be interpreted as necessarily representing the official policies, either expressed or implied, of NSF, ODNI, IARPA, or the U.S. Government. The U.S. Government is authorized to reproduce and distribute reprints for governmental purposes notwithstanding any copyright annotation therein.

\bibliography{custom}
\bibliographystyle{acl_natbib}

\clearpage

\appendix

\import{}{08_appendix.tex}

\end{document}

%% file: 01_introduction.tex

Text Simplification aims to improve the readability of texts with simpler grammar and word choices while preserving meaning \cite{Saiggon}. It provides reading assistance to children \cite{Kajiwara2013}, non-native speakers \cite{petersen2007text, pellow-eskenazi:2014:PITR, paetzold2016thesis}, and people with reading disabilities \cite{Rello:2013:ILS:2458308.2458354}. It also helps with downstream natural language processing tasks, such as parsing \cite{chandrasekar-etal-1996-motivations}, semantic role labelling \cite{vickrey-koller-2008-sentence}, information extraction \cite{miwa-etal-2010-entity}, and machine translation \citep[MT,][]{chen-etal-2012-simplification, stajner-popovic-2016-text}.

\setlength{\tabcolsep}{2.5pt}
\begin{table}[t!]
\centering
\small
\begin{tabular}{l|cccc}
  \hline
   & \textbf{OLen} & \textbf{\%new} & \textbf{\%eq} & \textbf{\%split}\\
  \hline
  Complex (input) & 20.7 & 0.0 & 100.0 & 0.0  \\  \hline
   \citet{narayan-gardent-2014-hybrid}$\dagger$  & 10.4 & 0.7 & 0.8 & 0.4\\ 
     \citeauthor{zhang-lapata-2017-sentence} \shortcite{zhang-lapata-2017-sentence}$\dagger$ & 13.8 & 8.1 & 16.8 & 0.0 \\
  \citeauthor{dong-etal-2019-editnts}
  \shortcite{dong-etal-2019-editnts}$\dagger$  & 10.9 & 8.4 & 4.6 & 0.0\\ 
  \citeauthor{kriz-etal-2019-complexity}
  \shortcite{kriz-etal-2019-complexity}$\dagger$ & 10.8 & 11.2 & 1.2 & 0.0\\\hline
   LSTM & 17.0 & 6.1 & 28.4 & 1.2 \\
  Our Model & 17.1 & 17.0 & 3.0 & 31.8 \\ \hline
   Simple (reference) & 17.9 & 29.0 & 0.0 & 30.0\\ \hline
   
\end{tabular}
\setlength{\belowcaptionskip}{-12pt}
\caption{Output statistics of 500 random sentences from the Newsela test set. Existing systems rely on deletion and do not paraphrase well. \textbf{OLen}, \textbf{\%new}, \textbf{\%eq} and \textbf{\%split} denote  the average output length, percentage of new words added, percentage of system outputs that are identical to the inputs, and percentage of sentence splits,  respectively. $\dagger$We used the system outputs shared by their authors.}
\label{table:intro_analysis}
\end{table}
\setlength{\tabcolsep}{4.0pt}

 Since 2016, nearly all text simplification systems have been sequence-to-sequence (seq2seq) models
 trained end-to-end, which have greatly increased the fluency of the outputs \cite{zhang-lapata-2017-sentence, nisioi-etal-2017-exploring, zhao-etal-2018-integrating,  kriz-etal-2019-complexity, dong-etal-2019-editnts, ACL-2020-chao}. However, these systems mostly rely on  deletion and tend to generate very short outputs at the cost of meaning preservation \cite{alva-manchego-etal-2017-learning}. 
 Table \ref{table:intro_analysis} shows that they neither split sentences nor paraphrase well as reflected by the low percentage of splits ($<$ 1\%) and new words introduced ($<$ 11.2\%).
 While deleting words is a viable (and the simplest) way to reduce the complexity of sentences, it is suboptimal and unsatisfying. Professional editors are known to use a sophisticated combination of deletion, paraphrasing, and sentence splitting to simplify texts \cite{Xu-EtAl:2015:TACL}.

\begin{figure*}
\centering
\includegraphics[width=0.95\linewidth]{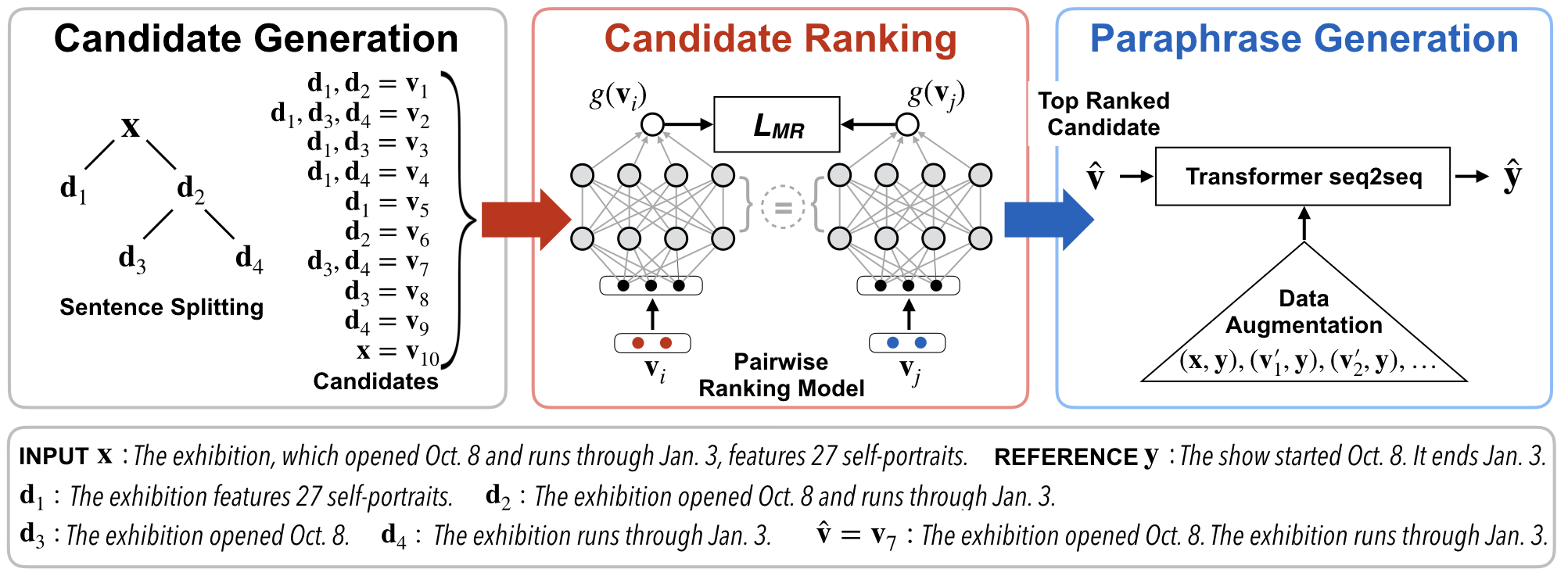}
\caption{Overview of our proposed model for text simplification, which can perform a controlled combination of sentence splitting, deletion, and paraphrasing. }
\label{fig:model_architecture}
\end{figure*}

Another drawback of these end-to-end neural systems is the lack of controllability. Simplification is highly audience dependant, and what constitutes simplified text for one group of users may not be acceptable for other groups \cite{Xu-EtAl:2015:TACL, lee-yeung-2018-personalizing}. An ideal simplification system should be able to generate text with varied characteristics, such as different lengths, readability levels, and number of split sentences, which can be difficult to control in end-to-end systems. 



To address these issues, we propose a novel hybrid approach that combines linguistically-motivated syntactic rules with data-driven neural models to improve  the diversity and controllability of the simplifications. We hypothesize that the seq2seq generation model will learn lexical and structural paraphrases more efficiently from the parallel corpus, when we offload some of the burden of sentence splitting (e.g., split at comma) and deletion (e.g., remove trailing preposition phrases) decisions to a separate component. 
Previous hybrid approaches for simplification~\citep{narayan-gardent-2014-hybrid,siddharthan-mandya-2014-hybrid, sulem-etal-2018-simple} used splitting and deletion rules in a deterministic step before applying an MT-based paraphrasing model. In contrast, our approach provides a more flexible and dynamic integration of linguistic rules with the neural models through ranking and data augmentation (Figure \ref{fig:model_architecture}).

We compare our method to several state-of-the-art systems in both automatic and human evaluations. Our model achieves overall better performance measured by SARI \cite{xu-etal-2016-optimizing} and other metrics, showing that the generated outputs are more similar to those written by human editors. We also demonstrate that our model can control the extent of each simplification operation by: (1) imposing a soft constraint on the percentage of words to be copied from the input in the seq2seq model, thus limiting lexical paraphrasing; and (2) selecting candidates that underwent a desired amount of splitting and/or deletion. Finally, we  create a new test dataset with multiple human references for Newsela \cite{Xu-EtAl:2015:TACL}, the widely used text simplification corpus, to specifically evaluate lexical paraphrasing. 

%% file: 03_model.tex
Figure \ref{fig:model_architecture} shows an overview of our hybrid approach. We  combine linguistic rules with data-driven neural models to improve the controllability and diversity of the outputs.
Given an input complex sentence $\mathbf{x}$, we first generate a set of intermediate simplifications $V = \{ \mathbf{v}_1, \mathbf{v}_2, \ldots, \mathbf{v}_n \}$ that have undergone splitting and deletion (\S \ref{sec:can_gen}). These intermediate sentences are then used for two purposes: (1) Selected by a pairwise neural ranking model (\S \ref{sec:ranking}) based on the simplification quality and then rewritten by the paraphrasing component; (2) Used for data augmentation to improve the diversity of the paraphrasing model (\S \ref{sec:paraphrase_generation}).

\subsection{Splitting and Deletion}
\label{sec:can_gen}

We leverage the state-of-the-art system for structural simplification, called \textbf{DisSim} \cite{niklaus-etal-2019-transforming}, to generate candidate simplifications that focus on splitting and deletion.\footnote{\url{https://github.com/Lambda-3/DiscourseSimplification}} The English version of DisSim applies 35 hand-crafted grammar rules to break down a complex sentence into a set of hierarchically organized sub-sentences (see Figure \ref{fig:model_architecture} for an example). 
We choose a rule-based approach for sentence splitting because it works really well. 
In our pilot experiments, DisSim successfully split 92\% of 100 complex sentences from the training data with more than 20 words, and introduced errors for only 6.8\% of these splits. 
We consider these sub-sentences as candidate simplifications for the later steps, except those that are extremely short or long (compression ratio $\notin$ [0.5, 1.5]). 
The compression ratio is calculated as the number of words in a candidate simplification $\mathbf{v}^i$ (which may contain one or more sub-sentences) divided by that of the original sentence $\mathbf{x}$. 

To further increase the variety of generated candidates, we supplement DisSim with a \textbf{Neural Deletion and Split} module trained on the text simplification corpus (\S \ref{sec:data}). We use a Transformer seq2seq model with the same configuration as the base model for paraphrasing (\S \ref{sec:paraphrase_generation}).  Given the input sentence $\mathbf{x}$, we constrain the beam search to generate 10 outputs with splitting and another 10 outputs without splitting. Then, we select the outputs that do not deviate substantially from $\mathbf{x}$ (i.e., Jaccard similarity $>$ 0.5). We add outputs from the two systems to the candidate pool $V$.




\subsection{Candidate Ranking}
\label{sec:ranking}

We design a neural ranking model to score all the candidates that underwent splitting and deletion, $V = \{ \mathbf{v}_1, \mathbf{v}_2, \ldots, \mathbf{v}_n \}$, then feed the top-ranked one to the lexical paraphrasing model for the final output. We train the model on a standard text simplification corpus consisting of pairs of complex sentence $\mathbf{x}$ and manually simplified reference $\mathbf{y}$. 

\paragraph{\textbf{Scoring Function.}} To assess the ``goodness'' of each candidate $\mathbf{v}_i$ during training, we define the $\textit{gold}$ scoring function $g^*$ as a length-penalized BERTscore:

\vspace{-6mm}
\begingroup
\begin{multline} \label{eq:bertscore}
    g^*(\mathbf{v}_i, \mathbf{y}) = e^{-\lambda|| \phi_{\mathbf{v}_i} -   \phi_{\mathbf{y}}||} \times 
    \\ BERTScore(\mathbf{v}_i, \mathbf{y})
\end{multline}
\endgroup
\vspace{-5mm}

\noindent BERTScore \cite{zhang2019bertscore} is a text similarity metric that uses BERT \cite{devlin-etal-2019-bert} embeddings to find soft matches between word pieces \cite{wu2016google} instead of exact string matching. We introduce a length penalty to favor the candidates that are of similar length to the human reference $\mathbf{y}$ and penalize those that deviate from the target compression ratio $\phi_{\mathbf{y}}$.   $\lambda$  defines the extent of penalization and is set to 1 in our experiments. $\phi_{\mathbf{v}_i}$ represents the compression ratios of $\mathbf{v}_i$ compared to the input $\mathbf{x}$. In principle, other similarity metrics can also be used for scoring.

\paragraph{\textbf{Pairwise Ranking Model.}} We train the ranking model in a pairwise setup since BERTScore is sensitive to the relative rather than absolute similarity, when comparing multiple candidates with the same reference. We transform the gold ranking of $V$ ($|V|=n$) into $n^2$ pairwise comparisons for every candidate pair, and learn to minimize the pairwise ranking violations using hinge loss:

\vspace{-3mm}
\begingroup
\begin{equation*}
\begin{split}
& L_{MR}=  \frac{1}{m} \sum_{k=1}^{m} \frac{1}{n^2_k } \sum_{i=1}^{n_k} \mathop{\sum_{j=1, i\neq j}^{{n_k}}} \max(0, 1 - l_{ij}^{k} d_{ij}^{k}) \\
& d_{ij}^{k} = g(\textbf{v}^{k}_{i})- g(\textbf{v}^{k}_{j}) \\
& l_{ij}^{k} = sign \left( g^*(\textbf{v}^{k}_{i}, \textbf{y}^k)- g^*(\textbf{v}^{k}_{j}, \textbf{y}^k) \right) \;\;\;\;\;\;\;\;\;\;\;\;\text{(2)} \\
\end{split}
\end{equation*}
\endgroup
\vspace{-3mm}

\noindent where $g(.)$ is a feedforward neural network, $m$ is the number of training complex-simple sentence pairs, $k$ is the index of training examples,
and $n_k$ represents the number of generated candidates (\S \ref{sec:can_gen}). On average, $n_k$ is about 14.5 for a sentence of 30 words, and can be larger for longer sentences. We consider 10 randomly sampled candidates for each complex sentence during training.

\paragraph{\textbf{Features.}} For the feedforward network $g(.)$, we use the following features: number of words in $\mathbf{v}_i$ and $\mathbf{x}$, compression ratio of $\textbf{v}_i$ with respect to $\mathbf{x}$, Jaccard similarity between $\mathbf{v}_i$ and $\mathbf{x}$, the rules applied on $\mathbf{x}$ to obtain $\mathbf{v}_i$, and the number of rule applications. We vectorize all the real-valued features using Gaussian binning \cite{EMNLP-2018-Maddela}, which has shown to help neural models trained on numerical features \cite{liu2016, sil2017neural, Zhong2020DiscourseLF}. We concatenate these vectors before feeding them to the ranking model. We score each candidate $\mathbf{v}_i$ separately and rank them in the decreasing order of $g(\mathbf{v}_i)$. We provide implementation details in Appendix \ref{app:implementation}.

\subsection{Paraphrase Generation}
\label{sec:paraphrase_generation}
We then paraphrase the top-ranked candidate $\hat{\mathbf{v}} \in V$ to generate the final simplification output $\mathbf{\hat{y}}$. Our paraphrase generation model can explicitly control the extent of lexical paraphrasing by specifying the percentage of words to be copied from the input sentence as a soft constraint. We also introduce a data augmentation method to encourage our model to generate more diverse outputs.

\paragraph{\textbf{Base Model.}} Our base generation model is a Transformer encoder-decoder initialized by the BERT checkpoint \cite{rothe-etal-2020-leveraging}, which achieved the best reported performance on text simplification in the recent work \cite{ACL-2020-chao}. We enhance this model with an attention-based copy mechanism to encourage  lexical paraphrasing, while remaining faithful to the input.

\paragraph{\textbf{Copy Control.}} Given the input candidate $\hat{\mathbf{v}} = (\hat{v}_1, \hat{v}_2, \ldots , \hat{v}_l)$ of $l$ words and the percentage of copying $cp \in (0,1]$, our goal is to paraphrase the rest of $(1-cp)\times l$ words in $\hat{\mathbf{v}}$ to a simpler version. To achieve this, we convert $cp$ into a vector of the same dimension as BERT embeddings using Gaussian binning \cite{EMNLP-2018-Maddela} and add it to the beginning of the input sequence $\hat{\mathbf{v}}$. The Transformer encoder then produces a sequence of context-aware hidden states $\mathbf{H} = (\mathbf{h}_1, \mathbf{h}_2 \ldots \mathbf{h}_l$), where $\mathbf{h}_i$ corresponds to the hidden state of $\hat{v}_i$. Each $\mathbf{h}_i$ is fed into the copy network which predicts the probability $p_i$ that word $\hat{v}_i$ should be copied to output. We create a new hidden state $\bar{\mathbf{h}}_i$ by adding $\mathbf{h}_i$ to a vector $\mathbf{u}$ scaled according to $p_i$. In other words, the scaled version of $\mathbf{u}$ informs the decoder whether the word should be copied. A single vector $\mathbf{u}$ is used across all sentences and hidden states, and is randomly initialized then updated during training. More formally, the encoding process can be described as follows:

\vspace{-3.8mm}
\begin{equation*}
\begin{aligned}
(\mathbf{h}_1, \mathbf{h}_2, \ldots , \mathbf{h}_l) & = encoder([cp; \hat{v}_1, \hat{v}_2, \ldots , \hat{v}_l])  \\
\bar{\mathbf{h}}_i & = \mathbf{h_i} + p_i \cdot \mathbf{u} ,  \\
\bar{\mathbf{H}} & = (\bar{\mathbf{h}}_1, \bar{\mathbf{h}}_2, \ldots , \bar{\mathbf{h}}_l)\;\;\;\;\;\;\;\;\;\;\;\;\;\;\text{(3)}\\ 
\end{aligned}
\end{equation*}
\vspace{-2.8mm}

\noindent The Transformer decoder generates the output sequence from $\bar{\mathbf{H}}$. Our copy mechanism is incorporated into the encoder rather than copying the input words during the decoding steps \cite{gu-etal-2016-incorporating, see-etal-2017-get}. Unless otherwise specified, we use the average copy ratio of the training dataset, 0.7, for our experiments.


\paragraph{\textbf{Multi-task Training.}} 

We train the paraphrasing model and the copy network in a multi-task learning setup, where predicting whether a word should be copied serves as an auxiliary task.  The gold labels for this task are obtained by checking if each word in the input sentence also appears in the human reference. When a word occurs multiple times in the input, we rely on the monolingual word alignment results from JacanaAlign \cite{yao-etal-2013-lightweight} to determine which occurrence is the one that gets copied. We train the Transformer model and the copy network jointly by minimizing the cross-entropy loss for both decoder generation and binary word classification.  We provide implementation and training details in Appendix \ref{app:implementation}.

\paragraph{\textbf{Data Augmentation.}} The sentence pairs in the training corpus often exhibit a variable mix of splitting and deletion operations along with paraphrasing (see Figure \ref{fig:model_architecture} for an example), which makes it difficult for the encoder-decoder models to learn paraphrases. Utilizing DisSim, we create additional training data that focuses on lexical paraphrasing 


For each sentence pair $\langle \mathbf{x}, \mathbf{y} \rangle$, we first generate a set of candidates $V = \{ \mathbf{v}_1, \mathbf{v}_2, \ldots, \mathbf{v}_n \}$ by applying DisSim to $\mathbf{x}$, as described in \S \ref{sec:can_gen}.  Then, we select a a subset of  $V$, called  $V' = \{ \mathbf{v}'_1, \mathbf{v}'_2, \ldots, \mathbf{v}'_{n'} \}$ ($V' \in V$) that are fairly close to the reference $\mathbf{y}$, but have only undergone splitting and deletion. We score each candidate $\mathbf{v}_i$ using the length-penalized BERTScore $g^*(\mathbf{v}_i, \mathbf{y})$ in Eq. (\ref{eq:bertscore}), and discard those with scores lower than 0.5. While calculating $g^*$, we set $\phi_{\mathbf{y}}$ and $\lambda$ to 1 and 2 respectively to favor candidates of similar length to the reference $\mathbf{y}$. We also discard the candidates that have different number of split sentences with respect to the reference. Finally, we train our model on the filtered candidate-reference sentence pairs $\langle \mathbf{v}'_1, \mathbf{y} \rangle$, $\langle \mathbf{v}'_2, \mathbf{y} \rangle$, $\ldots$ , $\langle \mathbf{v}'_{n'}, \mathbf{y} \rangle$, which focus on lexical paraphrasing, in addition to $\langle \mathbf{x}, \mathbf{y} \rangle$. 


\subsection{Controllable Generation}
\label{sec:con_gen}

We can control our model to concentrate on specific operations. For split- or delete-focused simplification, we select candidates with desirable length or number of splits during the candidate generation step. We perform only the paraphrase generation step for paraphrase-focused simplification. The paraphrasing model is designed specifically to paraphrase with minimal deletion and without splitting. It retains the length and the number of split sentences in the output, thus preserving the extent of deletion and splitting controlled in the previous steps. We control the degree of paraphrasing by changing the copy ratio.


%% file: 04_evaluation.tex
In this section, we compare our approach to various sentence simplification models using both automatic and manual evaluations. We show that our model achieves a new state-of-the-art and can adapt easily to different simplification styles, such as paraphrasing and splitting without deletion. 

\begin{table*}[ht!]
\small
\centering
\begin{tabular}{l|cccc|cc|ccc|ccc}
\hline
 \textbf{Models} & \textbf{SARI} & \textbf{add} & \textbf{keep}  & \textbf{del} & \textbf{FK} & \textbf{SLen} & \textbf{OLen} & \textbf{CR} & \textbf{\%split} & \textbf{s-BL} &  \textbf{\%new} & \textbf{\%eq}   \\ \hline 
Complex (input) & 15.9 & 0.0 & 47.6 & 0.0 & 12.0 & 23.7 & 23.8 & 1.0 & 0.0 & 100.0 & 0.0 & 100.0\\
Simple (reference) & 90.5 & 86.8 & 86.6 & 98.2 & 7.4 & 14.4 & 19.0 & 0.83 & 28.0 & 35.5 & 33.0 & 0.0   \\
\hline
LSTM & 35.0 & 1.6 & \textbf{45.5} & 57.8 & 8.9 & 17.6 & 17.9 & 0.8 & 1.9 & 66.5 & 5.0 & 20.2   \\
Hybrid-NG & 35.8 & 1.9 & 41.8 & 63.7 & 9.9 & 21.2 & 23.7 & 1.0 & 11.6 & 59.7 & 8.8 & 5.1  \\
Transformer$_{bert}$ &  37.0 & 3.1 & 43.6 & 64.4 & 8.1 & \textbf{15.6} & 20.2 & 0.87 & \textbf{24.1} & 58.8 & 12.8 & 10.2 \\
EditNTS & 38.1 & 1.6 & 45.8 & 66.5 & 8.5 & 16.0 & 21.4 & 0.92 & 32.0 & 71.4 & 8.3 & \textbf{0.2} \\
Our Model & \textbf{38.7} & \textbf{3.3} & 42.9 & \textbf{70.0} & \textbf{7.9} & 15.8 & \textbf{20.1} & \textbf{0.86} & 23.9 & \textbf{48.7} & \textbf{16.2} & 0.4 \\
\hline
\end{tabular}
\setlength{\belowcaptionskip}{-8pt}
\caption{Automatic evaluation results on \textsc{Newsela-Auto} test set. We report \textbf{SARI, the main automatic metric} for simplification, and its three edit scores namely precision for delete (\textbf{del}) and F1 scores for \textbf{add} and \textbf{keep} operations. We also report FKGL (\textbf{FK}), average sentence length  (\textbf{SLen}), output length (\textbf{OLen}), compression ratio (\textbf{CR}), self-BLEU (\textbf{s-BL}), percentage of sentence splits (\textbf{\%split}), average percentage of new words added to the output (\textbf{\%new}), and percentage of sentences identical to the input (\textbf{\%eq}). \textbf{Bold} typeface denotes the best performances (i.e., closest to the reference).}
\label{table:main_results}
\end{table*}

\subsection{Data and Experiment Setup}
\label{sec:data}

We train and evaluate our models on  Newsela  \cite{Xu-EtAl:2015:TACL}\footnote{\url{https://newsela.com/data/}} and Wikipedia copora \cite{zhu-etal-2010-monolingual, woodsend-lapata-2011-learning, coster-kauchak-2011-simple}. Newsela consists of 1,882 news articles with each article rewritten by professional editors for  students in different grades. We used the complex-simple sentence pairs automatically aligned by \citet{ACL-2020-chao}, called the {\sc Newsela-auto} dataset. 
 To capture sentence splitting, we joined the adjacent sentences in the simple article that are aligned to the same sentence in the complex article. Following \citet{stajner-etal-2015-deeper}, we removed the sentence pairs with high ($>$0.9) and low ($<$0.1) BLEU \cite{Papineni:2002:BMA:1073083.1073135} scores, which mostly correspond to the near identical and semantically divergent sentence pairs respectively. The final dataset consists of 259,778 train, 32,689 validation and 33,391 test complex-simple sentence pairs, where $\sim$30\% of pairs involve sentence splitting. Besides Newsela, we also provide the details of experiments on Wikipedia corpus in Appendix \ref{app:Wikipedia}, which show similar trends.

To demonstrate that our model can be controlled to generate diverse simplifications, we evaluate under the following settings: (i) Standard evaluation on the \textsc{Newsela-Auto} test set similar to the methodology in the recent literature \cite{ACL-2020-chao, dong-etal-2019-editnts, zhang-lapata-2017-sentence}, and (ii) Evaluation on different subsets of the \textsc{Newsela-Auto} test set that concentrate on a specific operation. We selected 9,356 sentence pairs with sentence splits for split-focused evaluation. Similarly, we chose 9,511 sentence pairs with compression ratio $<$ 0.7 and without sentences splits to evaluate delete-focused simplification. We created a new dataset, called  \textsc{Newsela-Turk}, to evaluate lexical paraphrasing.\footnote{We also provide results on 8,371 sentence pairs of \textsc{Newsela-Auto} test set with compression ratio $>$ 0.9 and no splits in Appendix \ref{app:paraphrase_eval}, which show similar trends.}
Similar to the \textsc{Wikipedia-Turk} benchmark corpus \cite{xu-etal-2016-optimizing}, \textsc{Newsela-Turk} consists of human-written references focused on lexical paraphrasing.
We first selected sentence pairs from the \textsc{Newsela-Auto} test set of roughly similar length (compression ratio between 0.8 and 1.2) and no sentence splits because they more likely involve paraphrasing. Then, we asked Amazon Mechanical Turk workers to simplify the complex sentence without any loss in meaning.\footnote{We provide instructions in Appendix \ref{app:ann_interface}} To ensure the quality of simplifications, we manually selected the workers using the qualification test proposed in \citet{fern2020asset}, during which the workers were asked to simplify three sentences. We selected top 35\% of the 300 workers that participated in the test. We periodically checked the submissions and removed the bad workers. In the end, we collected 500 sentences with 4 references for each sentence.

\setlength{\tabcolsep}{3pt}
\begin{table*}[ht!]
\small
\centering
\begin{tabular}{l|cccc|cc|ccc|ccc}
\hline
 \textbf{Models} & \textbf{SARI} & \textbf{add} & \textbf{keep}  & \textbf{del} & \textbf{FK} & \textbf{SLen} & \textbf{OLen} & \textbf{CR} & \textbf{\%split} & \textbf{s-BL} &  \textbf{\%new} & \textbf{\%eq}  \\ \hline 

 Complex (input) &  22.3 & 0.0 & 67.0 & 0.0 & 12.8 & 23.3 & 23.5 & 1.0 & 0.0 & 100.0 & 0.0 & 100.0 \\
 Simple (reference)&  62.3 & 44.8 & 68.3 & 73.9 & 11.1 & 23.8 & 23.5 & 1.01 & 0.0 & 48.5 & 24.1 & 0.0 \\
\hline
Hybrid-NG & 38.2 & 2.8 & 57.0 & 54.8 & 10.7 & 21.6 & 23.1 & 0.98 & 7.0 & 57.2 & 9.1 & 1.4\\
Transformer$_{bert}$ & 36.0 & 3.3 & 54.9 & 49.8 & 8.9 & 16.1 & 20.2 & 0.87 & 23.0 & 58.7 & 13.3 & 7.6 \\
EditNTS & 37.4 & 1.6 & 61.0 & 49.6 & 9.5 & 16.9 & 21.9 & 0.94 & 0.0 & 73.1 & 5.8 & \textbf{0.0} \\
\hline
Our Model & 
38.1 & \textbf{3.9} & 55.1 & 55.5 & 8.8 & 16.6 & 20.2 & 0.86 & 19.6 & \textbf{50.4} & 15.7 & \textbf{0.0} \\
Our Model (no split; $cp$ = 0.6) & 
39.0 & 3.8 & 57.7 & 55.6 & \textbf{11.2} & 22.1 & 22.9 & 0.98 & 0.2 & 55.9 & \textbf{18.0} & 1.0   \\
Our Model (no split; $cp$ = 0.7) & 
 \textbf{41.0} & 3.4 & 63.1 & \textbf{56.6} & 11.5 & 22.2 & 22.9 & 0.98 & \textbf{0.0} & 69.4 & 10.4 & 4.2\\
Our Model (no split; $cp$ = 0.8) & 40.6 & 2.9 & \textbf{65.0} & 54.0 & 11.8 & \textbf{22.4} & \textbf{23.2} & \textbf{0.99} & \textbf{0.0} & 77.7 & 6.6 & 10.8  \\
\hline
\end{tabular}
\caption{Automatic evaluation results on \textsc{Newsela-Turk} that focuses on paraphrasing (500 complex sentences with 4 human written paraphrases). We control the extent of paraphrasing of our models by specifying the percentage of words to be copied ($cp$) from the input as a soft constraint.}
\label{table:paraphrase_results}
\end{table*}

\begin{table*}[ht!]
\small
\centering
\begin{tabular}{p{3.8cm}|cccc|cc|ccc|ccc}
\hline
 \textbf{Models} & \textbf{SARI} & \textbf{add} & \textbf{keep}  & \textbf{del} & \textbf{FK} & \textbf{SLen} & \textbf{OLen} & \textbf{CR} & \textbf{\%split} & \textbf{s-BL} &  \textbf{\%new} & \textbf{\%eq}  \\ \hline 
Complex (input) &  17.0 & 0.0 & 51.1 & 0.0 & 14.6 & 30.0 & 30.2 & 1.0 & 0.0 & 100.0 & 0.0 & 100.0 \\
Simple (reference) &  93.0 & 89.9 & 91.6 & 97.5 & 7.0 & 13.4 & 28.6 & 0.98 & 100.0 & 36.8 & 29.7 & 0.0 \\
\hline
Hybrid-NG & 37.1 & 2.2 & 44.9 & 64.1 & 11.6 & 25.5 & \textbf{30.1} & \textbf{1.0} & 17.3 & 57.7 & 8.7 & 1.6 \\
Transformer$_{bert}$ & 39.5 & 4.2 & 47.3 & 67.0 & 8.8 & 17.1 & 25.3 & 0.85 & 39.7 & 57.7 & 11.9 & 5.2  \\
EditNTS & 38.9 & 1.5 & 49.1 & 66.2 & 9.1 & 16.9 & 26.2 & 0.88 & 50.3 & 71.2 & 7.2 & 0.2 \\
\hline
Our Model & 
  39.4 & 4.0 & 46.6 & 67.6 & 8.7 & 17.5 & 25.5 & 0.85 & 40.6 & \textbf{48.3} & \textbf{15.6} & \textbf{0.1}  \\
Our Model (w/ split) &  \textbf{42.1} & \textbf{5.6} & \textbf{50.6} & \textbf{70.1} & \textbf{8.1} & \textbf{15.3} & 30.3 & 1.02 & \textbf{93.5} & 60.7 & 12.4 & 1.1 \\
\hline
\end{tabular}
\caption{Automatic evaluation results on a splitting-focused subset of the \textsc{Newsela-Auto} test set (9,356 sentence pairs with splitting).  Our model chooses only candidates that have undergone splitting during the ranking step.}
\label{table:split_results}
\end{table*}

\begin{table*}[ht!]
\small
\centering
\begin{tabular}{p{4.0cm}|cccc|cc|ccc|ccc}
\hline
 \textbf{Models} & \textbf{SARI} & \textbf{add} & \textbf{keep}  & \textbf{del} & \textbf{FK} & \textbf{SLen} & \textbf{OLen} & \textbf{CR} & \textbf{\%split} & \textbf{s-BL} &  \textbf{\%new} & \textbf{\%eq}  \\ \hline 
Complex (input) &   9.6 & 0.0 & 28.8 & 0.0 & 12.9 & 25.8 & 26.0 & 1.0 & 0.0 & 100.0 & 0.0 & 100.0 \\
Simple (reference) &  85.7 & 82.7 & 76.0 & 98.6 & 6.7 & 12.6 & 12.6 & 0.5 & 0.0 & 19.6 & 32.6 & 0.0 \\
\hline
Hybrid-NG & 35.8 & 1.4 & 27.0 & 79.1 & 10.6 & 22.7 & 25.9 & 1.0 & 13.3 & 58.9 & 8.7 & 3.6  \\
Transformer$_{bert}$ & 36.8 & 2.2 & 29.6 & 78.7 & 8.4 & \textbf{16.2} & 21.7 & 0.85 & 27.7 & 57.9 & 12.3 & 8.2  \\
EditNTS &   37.1 & 1.0 & 29.7 & 80.7 & 8.8 & 16.6 & 23.1 & 0.91 & 36.6 & 71.8 & 7.8 & 0.6 \\
\hline
Our Model & 
 \textbf{39.2} & \textbf{2.4} & \textbf{29.8} & \textbf{85.3} & \textbf{8.2} & 16.4 & 21.9 & 0.85 & 29.1 & 48.8 & \textbf{15.6} & 0.4 \\
Our Model (no split; CR$<$0.7) & 38.2 & 2.0 & 28.5 & 84.1 & 8.6 & 16.8 & \textbf{17.5} & \textbf{0.68} & \textbf{0.1} & \textbf{42.0} & 12.5 & \textbf{0.2} \\
\hline
\end{tabular}
\setlength{\belowcaptionskip}{-6pt}
\caption{Automatic evaluation results on a deletion-focused subset of the \textsc{Newsela-Auto} test set (9,511 sentence pairs with compression ratio $<$ 0.7 and no sentence splits). Our model selects only candidates with similar compression ratio and no splits during ranking.}
\label{table:comp_results}
\end{table*} 

\subsection{Existing Methods}
We use the following simplification approaches as baselines:
(i) \textbf{BERT-Initialized Transfomer} \cite{rothe-etal-2020-leveraging}, where the encoder is initialized with BERT$_{base}$ checkpoint and the decoder is randomly initialized. It is the current state-of-the-art for text simplification \cite{ACL-2020-chao}. 
(ii) \textbf{EditNTS} \cite{dong-etal-2019-editnts},\footnote{\url{https://github.com/yuedongP/EditNTS}} another state-of-the-art model that uses a neural programmer-interpreter \cite{reed-defreitas:2016} to predict the edit operation on each word, and then generates the simplified sentence.
(iii) \textbf{LSTM baseline}, a vanilla encoder-decoder model used in \citeauthor{zhang-lapata-2017-sentence} \shortcite{zhang-lapata-2017-sentence}.
(iv) \textbf{Hybrid-NG} \cite{narayan-gardent-2014-hybrid},\footnote{\url{https://github.com/shashiongithub/Sentence-Simplification-ACL14}} one of the best existing hybrid systems that performs splitting and deletion using a probabilistic model and lexical substitution with a phrase-based machine translation system. We retrained all the models on the \textsc{Newsla-Auto}  dataset.

\subsection{Automatic Evaluation} 
\label{sec:automatic_eval}

\import{}{05_controlled_generation.tex}

\subsection{Human Evaluation}
We performed two human evaluations: one to measure the overall simplification quality and the other to specifically capture sentence splitting.\footnote{We provide instructions in Appendix \ref{appendix:human_eval}.} For the first one, we asked five Amazon Mechanical Turk workers to evaluate fluency, adequacy and simplicity of 100 random simplifications from the \textsc{Newsela-Auto} test set.
We supplemented the fluency and adequacy ratings with binary questions described in \citet{zhang2020small} for the second evaluation over another 100 simplifications from the \textsc{Newsela-Auto} split-focused test set. We asked if the output sentence exhibits spitting and if the splitting occurs at the correct place.
While fluency measures the grammaticality of the output, adequacy captures the extent of meaning preserved when compared to the input. Simplicity  evaluates if the output is simpler than the input. Each sentence was rated on a 5-point Likert scale and we averaged the ratings from the five workers. We chose the majority value for the binary ratings. We used the output of our model that is tailored for sentence splitting for the second evaluation. 

Table \ref{table:human_eval} demonstrates that our model achieves the best fluency, simplicity, and overall ratings. The adequacy rating is also very close to that of Transformer$_{bert}$ and EditNTS even though our model is performing more paraphrasing (Table \ref{table:main_results}), which verifies that the changes made by our system are meaningful. Our model  achieves the most number of correct sentence splits (90\%), and the highest fluency (4.19) for syntactic simplification, showing that it can generate more number of coherent sentence splits when compared to other models.

\section{Model Analysis}
In this section, we analyze the contribution of each model component  and examine the system errors. 

\begin{table*}[h!]
\centering
\small
\begin{tabular}{p{1.8cm}|p{13cm}}
\hline
&\textbf{Examples} \\
  \hline
\multicolumn{2}{l}{\textbf{Good} (49\%) } \\
\hline
 Complex & \textit{The Seattle kids petitioned Washington state last year to adopt stricter science-based regulations to protect them against climate change.} \\
Simple & \textit{The eight kids from Seattle who petitioned Washington state last year are fighting for the state to adopt stricter laws to protect their future against climate change.} \\
Our Model & \textit{The Seattle kids {\color{blue} \textbf{asked}} the Washington state last year {\color{blue} \textbf{to take tougher rules against}} climate change.}  \\
\hline
\multicolumn{2}{l}{\textbf{Hallucinations} (13\%) } \\
\hline
Complex & \textit{``It's more of a family than living outside,'' said Jessica Konczal, 33, whose husband is Sgt. Matthew Konczal .} \\
Simple & \textit{Jessica Konczal, age 33, says the base is its own type of family .}  \\
Our Model & \textit{``It's more of a family than living outside,'' said Jessica Konczal, {\color{red} \textbf{one of the protesters}}.}  \\
\hline
\multicolumn{2}{l}{\textbf{Fluency Errors} (14\%) } \\
\hline
Complex & \textit{Parental feedback on the menu additions so far, from some of the early adopter markets, has been ``phenomenal,'' Leverton said.} \\
Simple & \textit{Leverton said parents love the new food.} \\
Our Model & \textit{So far, {\color{red} \textbf{parents parents have feedback }} on the menu {\color{blue} \textbf{changes}} has been {\color{blue} \textbf{a great deal}}, {\color{red} \textbf{he}} {\color{blue} \textbf{added}}.}  \\
\hline
\multicolumn{2}{l}{\textbf{Bad Substitution} (15\%) } \\
\hline
Complex & \textit{One of the device's inventors explained to the president that the machine was a prototype.} \\
Simple & \textit{One of the girls explained it was just a prototype.}  \\
Our Model & \textit{One of the inventors {\color{blue} \textbf{told}} the president that the {\color{red} \textbf{first design was a design}}.}  \\
\hline
\multicolumn{2}{l}{\textbf{Anaphora Resolution / Entailment} (12\%) } \\
\hline
Complex & \textit{Sea slugs dubbed sacoglossans are some of the most remarkable biological burglars on the planet.} \\
Simple & \textit{The sea slugs known as sacoglossans are some of the most interesting burglars on the planet.} \\
Our Model & \textit{Scientists say {\color{red} \textbf{these}} are some of the most {\color{blue} \textbf{interesting}} {\color{red} \textbf{creatures}} on the planet.}  \\
\hline
\multicolumn{2}{l}{\textbf{Human Reference Errors} (4\%) } \\
\hline
Complex & \textit{They float in and out of formations that combine the underwater world with the stage. 
} \\
Simple & \textit{In the dance on a round stage on the seabed, the dancers swim together and apart.}\\
Our Model & \textit{They float in and out of {\color{blue} \textbf{places}} that combine stage with the underwater.}  \\
\hline
\end{tabular}
\setlength{\belowcaptionskip}{-8pt}
\caption{Error analysis of 200 simplifications  from \textsc{Newsela-Auto} test set generated by our model. {\color{red} \textbf{Red}} marks the errors; {\color{blue}\textbf{blue}} marks good paraphrases.}
\label{table:error_analysis}
\end{table*}

\begin{table}[t!]
\small
\centering
\begin{tabular}{p{2.5cm}|ccccc}
\hline
& \textbf{SARI} & \textbf{FK} & \textbf{CR}  & \textbf{\%split} & \textbf{\%new} \\ 
\hline 
Complex (input) & 15.9 & 12.2 & 1.0 & 0.0 & 0.0  \\
Simple (reference) & 90.5 & 7.5 & 0.83 & 28.9 & 32.8 \\   
\hline
Random Candidate & 33.7 & 8.1 & 0.81 & 34.4 & 5.5 \\
BERTScore$_{len}$ &  36.9 & 9.0 & 0.87 & 25.9 & 5.9  \\
\hline
Our Model & \textbf{38.6} & 8.4 & 0.88 & 26.1 & 
\textbf{18.9}  \\
$-$ augmentation &  37.6 & 7.9 & 0.86 & \textbf{29.5} & 12.7 \\
$-$ copy attn & 36.0 & 8.1 & 0.87 & 26.2 & 15.9 \\
$-$ only Transformer&  37.9 & \textbf{7.7} & 0.78 & 26.3 & 16.5  \\
$-$ only DisSim&  37.2 & 8.3 & \textbf{0.84} & 27.1 & 18.0  \\
\hline
\end{tabular} 
\setlength{\belowcaptionskip}{-6pt}
\caption{Model ablation study on dev set}
\label{table:ablation}
\end{table}

\subsection{System Ablations}
We evaluate our key design choices, namely candidate ranking that is based on length-penalized BERTScore and paraphrase generation that uses data augmentation and copy attention. Table \ref{table:ablation} summarizes the results. Our pairwise ranking model (BERTScore$_{len}$) achieves an increase of 3.2 points in SARI when compared to choosing a random (Random) candidate. Randomly selecting a candidate also performs fairly well, indicating that the sentence splitting and deletion models we chose are of good quality.

Compared to our final model (Our Model), its variants without data augmentation ($-$ augmentation) and copy mechanism ($-$ copy attn) suffer a drop of 1.0 and 2.6 points in SARI respectively and a decrease of at least 3.0\% of new words, which demonstrates that these components encourage the system to paraphrase. Our model trained on only DisSim ($-$ only DisSim) and Transformer ($-$ only Transformer) candidates  performs close to our best model (Our Model) in terms of SARI. 

\subsection{Error Analysis}

To understand the errors generated by our model, we manually classified 200 simplifications from the \textsc{Newsela-Auto} test set  into the following categories: (a) \textbf{Good}, where the model generated meaningful simplifications, (b) \textbf{Hallucinations}, where the model introduced information not in the input, (c) \textbf{Fluency Errors}, where the model generated ungrammatical output, (d) \textbf{Anaphora Resolution}, where it was difficult to resolve pronouns in the output. (e) \textbf{Bad substitution}, where the model inserted an incorrect simpler phrase, and (e) \textbf{Human Reference Errors}, where the reference does not reflect the source sentence. Note that a simplification can belong to multiple error categories. Table \ref{table:error_analysis} shows the examples of each category.

%% file: 05_controlled_generation.tex
\noindent \paragraph{Metrics.}  We report \textbf{SARI} \cite{xu-etal-2016-optimizing},
 which averages the F1/precision of n-grams $(n \in \{1, 2, 3, 4\})$ inserted, deleted and kept when compared to human references. More specifically, it computes the F1 score for the n-grams that are added (\textbf{add}),\footnote{We slightly improved the SARI implementation by \citet{xu-etal-2016-optimizing} to exclude the spurious ngrams while calculating the F1 score for \textbf{add}. For example, if the input contains the phrase ``\textit{is very beautiful}'', the phrase ``\textit{is beautiful}'' is treated as a new phrase in the original implementation even though it is caused by the delete operation.}  which is an important indicator if a model is good at paraphrasing. The model's deletion capability is measured by the F1 score for n-grams that are kept (\textbf{keep}) and precision for those deleted (\textbf{del}).\footnote{SARI score of a reference with itself may not always be 100 as it considers 0 divided by 0 as 0, instead of 1, when calculating n-gram precision and recall. This avoids the inflation of \textbf{del} scores when the input is same as the output.} To evaluate a model's paraphrasing capability and diversity, we calculate the BLEU score with respect to the input (\textbf{s-BL}), the percentage of new words (\textbf{\%new}) added, and the percentage of system outputs identical to the input (\textbf{\%eq}). Low s-BL, \%eq, or high \%new indicate that the system is less conservative. 
We also report Flesch-Kincaid (\textbf{FK}) grade level readability \cite{kincaid}, average sentence length (\textbf{SLen}), the percentage of splits (\textbf{\%split}), compression ratio (\textbf{CR}), and average output length (\textbf{OLen}). We do not report BLEU because it often does not correlate with simplicity \cite{sulem-etal-2018-bleu, sulem-etal-2018-semantic, xu-etal-2016-optimizing}.


\begin{table*}[t!]
\small
\centering
\begin{tabular}{l|cccc|cccc}
\hline
& \multicolumn{4}{c|}{\textbf{Overall Simplification}} & \multicolumn{4}{c}{\textbf{Structural Simplification}}\\
\textbf{Model} & \textbf{Fluency} & \textbf{Adequacy} & \textbf{Simplicity} & \textbf{Average} & \textbf{Fluency} & \textbf{Adequacy} & \textbf{Has Split} & \textbf{Correct Split} \\ \hline 
Hybrid-NG  & 3.23* & 2.96* & 3.40* & 3.19* & 3.25* & 3.53* & 42\% & 15\%\\
EditNTS  & 3.88* & 
\textbf{3.70} & 3.67 & 3.75 &  4.08 & 3.81* & 41\% & 18\% \\
Transformer$_{bert}$ & 3.91 & 3.63 & 3.65* & 3.73 & 4.15 & 3.65* & 53\% & 49\%  \\
Our Model &  \textbf{4.02} & 3.65 & \textbf{3.77} & \textbf{3.81}  & 4.19 & 4.13 & 97\% & 90\% \\
\hline
Simple (reference) & 4.12 & 3.64 & 3.97 & 3.84 & 4.41 & 4.10 & 100\% & 100\%  \\
\hline
\end{tabular}
\setlength{\belowcaptionskip}{-6pt}
\caption{Human evaluation of 100 random simplifications from the \textsc{Newsela-Auto} test set and the split-focused subset of the same test set. \textbf{Has Split} and \textbf{Correct Split} denote the percentage of the output sentences that have undergone splitting and the percentage of coherent splits respectively. * denotes that our model is significantly better than the corresponding baseline  (according to a $t$-test with $p < 0.05$).}
\label{table:human_eval}
\end{table*} 

\vspace{-5pt}
\noindent \paragraph{Results.}  Table \ref{table:main_results} shows the results on \textsc{Newsela-Auto} test set. Our model outperforms the state-of-the-art  Transformer$_{bert}$ and EditNTS models with respect to SARI.\footnote{According to \citet{ACL-2020-chao}, a BERT-initialized Transformer performs better than EditNTS. We see a different behavior here because we retained sentence splits from 0-1, 1-2, 2-3 readability levels in {\sc Newsela-Auto}, which contained more lexical overlaps and inflated the scores for EditNTS.} EditNTS and LSTM focus on deletion as they show high self-BLEU ($>$66.5) and FK ($>$8.8) scores despite having compression ratios similar to other systems. Transformer model alone is rather conservative and copies 10.2\% of the sentences directly to the output. Although Hybrid-NG makes more changes than any other baselines, its SARI and add scores are 3.7 and 1.7 points lower than our model indicating that it generates more errors. Our model achieves the lowest self-BLEU (48.7), FK (7.9), and percentage of sentences identical to the input (0.4), and the highest add (3.3) score and percentage of new words (16.2\%). In other words, our system is the least conservative, generates more good paraphrases, and mimics the human references better. We provide examples of system outputs in Table \ref{table:sys_out} and Appendix \ref{app:sys_out}. 

Tables \ref{table:paraphrase_results}, 
 \ref{table:split_results}, and \ref{table:comp_results} show the results on \textsc{Newsela-Turk}, split-focused, and delete-focused subsets of \textsc{Newsela-Auto} test set respectively. For these experiments, we configure our model to focus on specific operations (details in \S \ref{sec:con_gen}). Our model again outperforms the existing systems according to SARI, add score, and percentage of new words, which means that our model is performing more meaningful paraphrasing. We show that we can control the extent of paraphrasing by varying the copy ratio ($cp$). Our model splits 93.5\% of the sentences, which is substantially better than the other models.

%% file: 06_related_work.tex
\begin{table*}[t!]
\centering
\small
\begin{tabular}{p{2.7cm}|p{12.8cm}}

  \hline
  &\textbf{System Outputs} \\
  \hline
Complex   &   \textit{Since 2010, project researchers have uncovered documents in Portugal that have revealed who owned the ship.} \\
Simple   &   \textit{Since 2010, experts have been figuring out who owned the ship.} \\
\hline
Hybrid-NG   &   \textit{since 2010, the project {\color{blue}\textbf{scientists}} have uncovered documents in portugal that have {\color{blue}\textbf{about}} who {\color{blue}\textbf{owns}} the ship.} \\
LSTM   &   \textit{since 2010, {\color{blue}\textbf{scientists}} have uncovered documents in portugal that have revealed who owned the ship.} \\
Transformer$_{bert}$  &   \textit{{\color{red}\textbf{they}} {\color{blue}\textbf{discovered}} that the ship {\color{red}\textbf{had been important}}.} \\
EditNTS   &   \textit{since 2010, project researchers have uncovered documents in portugal{\color{red}\textbf{. have}} revealed who owned the ship.} \\
\hline
Our Model ($cp$ = 0.6)   &   \textit{{\color{blue}\textbf{scientists}} have {\color{blue}\textbf{found}} a {\color{red}\textbf{secret deal}}{\color{blue}\textbf{. they have discovered}} who owned the ship.} \\
Our Model ($cp$ = 0.7)   &   \textit{{\color{blue}\textbf{scientists}} have {\color{blue}\textbf{found}} documents in portugal{\color{blue}\textbf{. they have also found out}} who owned the ship.} \\
Our Model ($cp$ = 0.8)   &   \textit{{\color{blue}\textbf{scientists}} have {\color{blue}\textbf{found}} documents in portugal{\color{blue}\textbf{. they have discovered}} who owned the ship.} \\
\hline
\hline
Complex   &   \textit{Experts say China's air pollution exacts a tremendous toll on human health.} \\
Simple   &   \textit{China's air pollution is very unhealthy.} \\
\hline
Hybrid-NG   &   \textit{experts say the {\color{red}\textbf{government's}} air pollution exacts a toll on human health.} \\
LSTM   &   \textit{experts say china's air pollution exacts a tremendous toll on human health.} \\
Transformer$_{bert}$  &   \textit{experts say china's pollution has a tremendous {\color{blue}\textbf{effect}} on human health.} \\
EditNTS   &   \textit{experts say china's air pollution {\color{red}\textbf{can cause}} human health.} \\
\hline
Our Model ($cp$ = 0.6)   &   \textit{experts say china's air pollution {\color{blue}\textbf{is a big problem}} for human health.} \\
Our Model ($cp$ = 0.7)   &   \textit{experts say china 's air pollution {\color{blue}\textbf{can cause a lot of damage}} on human health.} \\
Our Model ($cp$ = 0.8)   &   \textit{experts say china 's air pollution {\color{blue}\textbf{is}} a {\color{blue}\textbf{huge}} toll on human health.} \\
\hline
\end{tabular}
\caption{Examples of system outputs. {\color{red} \textbf{Red}} marks the errors; {\color{blue}\textbf{blue}} marks good paraphrases.  $cp$ is a soft constraint that denotes the percentage of words that can be copied from the input.}
\label{table:sys_out}
\end{table*}

Before the advent of neural networks, text simplification approaches performed each operation separately in a pipeline manner using either handcrafted rules \cite{E99-1042, siddharthan-pipeline, siddharthan-etal-2004-syntactic} or data-driven methods based on parallel corpora \cite{zhu-etal-2010-monolingual,  woodsend-lapata-2011-learning, narayan-gardent-2014-hybrid}. Following neural machine translation, the trend changed to performing all the operations together end-to-end \cite{zhang-lapata-2017-sentence, nisioi-etal-2017-exploring, zhao-etal-2018-integrating, alva-manchego-etal-2017-learning,
vu-etal-2018-sentence,
kriz-etal-2019-complexity, dong-etal-2019-editnts, ACL-2020-chao} at the cost of controllability and performance as shown in this paper. 

Controllable text simplification has been attempted before, but only with limited capability. \citet{scarton-specia-2018-learning} and \citet{martin2019controllable} added additional tokens to the input representing grade level, length, lexical, and structural complexity constraints. \citet{nishihara-etal-2019-controllable} proposed a loss which controls word complexity, while \citet{mallinson2019controllable} concatenated constraints to each word embedding. \citet{dhruv-acl-2020} proposed a linguistic scoring function to control the edits to the input.

Another long body of research focuses on a single simplification operation and can be broadly divided into three categories: (1) Lexical Simplification \cite{specia2012, horn-manduca-kauchak:2014:P14-2, glavavs-vstajner:2015:ACL-IJCNLP, paetzold2017a, paetzold-specia:2015:ACL-IJCNLP-2015-System-Demonstrations, EMNLP-2018-Maddela, qiang2019BERTLS}, where complex words are substituted with simpler words. (2) Syntactic Simplification \cite{siddharthan2006syntactic, aharoni-goldberg-2018-split, botha-etal-2018-learning, niklaus-etal-2019-transforming}, which deals exclusively with sentence  splitting, and (3) Sentence Compression \cite{filipova2015, rush-etal-2015-neural, nallapati-etal-2016-abstractive,see-etal-2017-get, baziotis-etal-2019-seq}, where the goal is to shorten the input sentence by removing its irrelevant content.

%% file: 07_conclusion.tex
 We proposed a novel hybrid approach for sentence simplification that performs better and produces more diverse outputs than the existing systems.  We designed a new data augmentation method to encourage the model to paraphrase. We created a new dataset, \textsc{Newsela-Turk}, to evaluate paraphrasing-focused simplifications. We showed that our model can control various attributes of the simplified text, such as number of sentence splits, length, and number of words copied from the input.

%% file: 08_appendix.tex
\section{Implementation and Training Details}
\label{app:implementation}

We implemented two separate Transformer models for neural deletion and split component (\S \ref{sec:can_gen}) and paraphrase generation (\S \ref{sec:paraphrase_generation}) using the Fairseq\footnote{\url{https://github.com/pytorch/fairseq}} toolkit. Both the encoder and decoder follow BERT$_{base}$\footnote{\url{https://github.com/google-research/bert}} architecture, while the encoder is also initialized with BERT$_{base}$ checkpoint. For neural deletion  and split component, we used a beam search of width 10 to generate candidates. The copy attention mechanism is a feedforward network containing 3 hidden layers, 1000 nodes in each layer with \textit{tanh} activation, and a single linear output node  with \textit{sigmoid} activation. We used Adam optimizer \cite{Kingma2014} with a learning rate of 0.0001, linear learning rate warmup of 40k steps, and 100k training steps. We used a batch size of 64. We used BERT WordPiece tokenizer. During inference, we constrained the beam-search to not repeat trigrams and emitted sentences that avoided aggressive deletion (compression ratio $\in [0.9, 1.2]$. We chose the best checkpoint based on the SARI score \cite{xu-etal-2016-optimizing} on the dev set. We saved a checkpoint after every epoch. We did not perform any hyperparameter search and directly used the hyperparameters of the BERT-initialized Transformer described in \citet{rothe-etal-2020-leveraging}. The model takes 10 hours to train on 1 NVIDIA GeForce GPU.

Our pairwise ranking model, implemented using the PyTorch framework, consists of 3 hidden layers, 100 nodes in each layer, $tanh$ activation, and a single linear output node. We used Adam optimizer with a learning rate of 0.01 and 10 epochs. We applied a dropout of 0.2. For Gaussian binning, we vectorized the numerical features into 10 dimensional vectors. The model takes half hour to train on 1 NVIDIA GeForce GPU. We do not perform any extensive hyperparameter tuning. We just examined few values for learning rate (0.001, 0.01 and 0.1) and chose the best based on the SARI score on the dev set. We used the original code for DisSim.\footnote{\url{https://github.com/Lambda-3/DiscourseSimplification}}





\onecolumn
\section{Annotation Interface}
\label{app:ann_interface}
\begin{minipage}{\textwidth}
    \centering
    \includegraphics[width=0.95\textwidth]{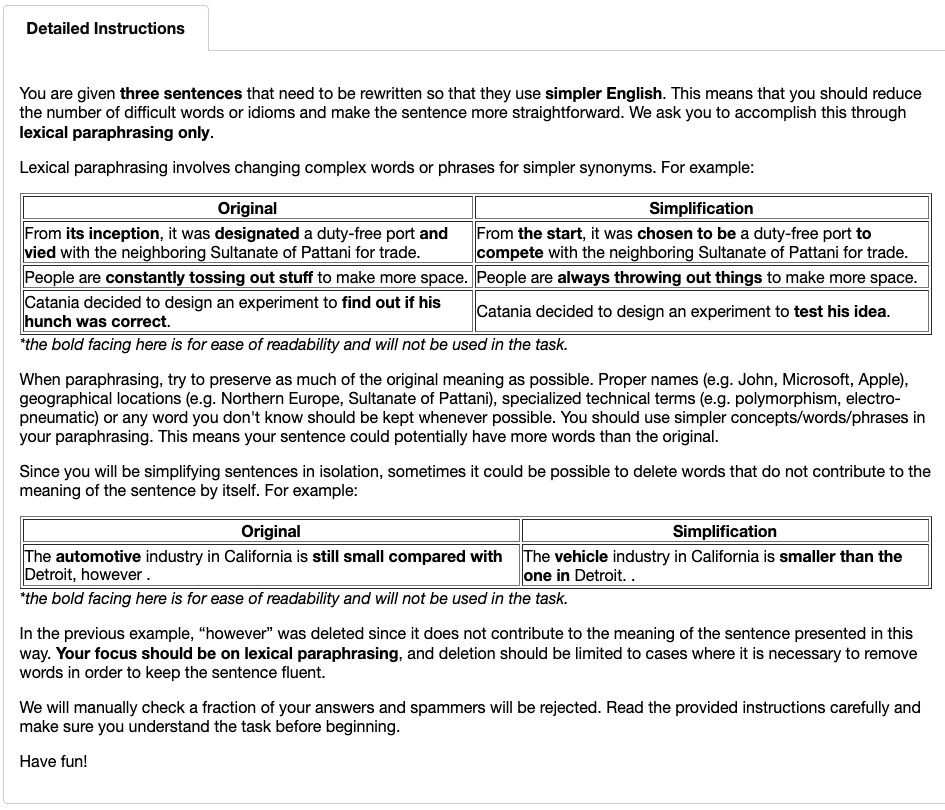}
    \includegraphics[width=0.95\textwidth]{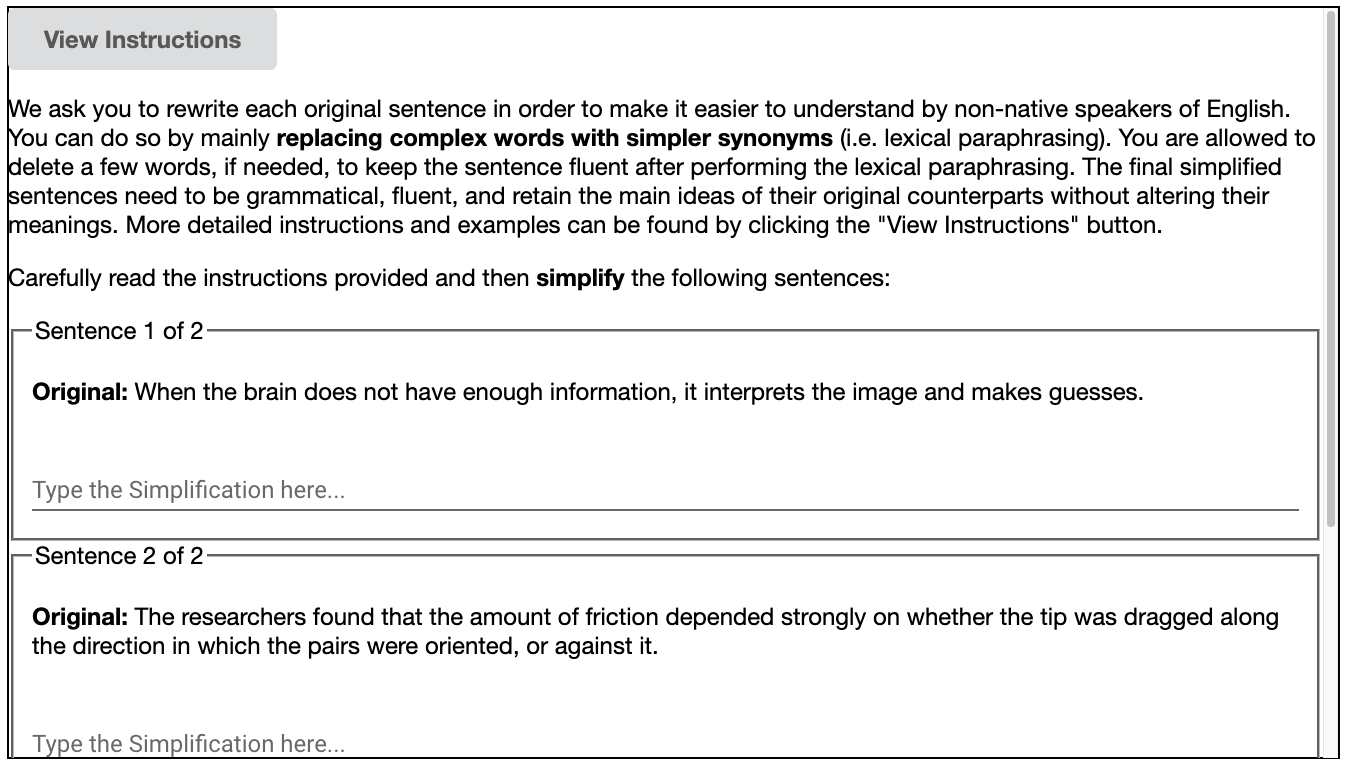}
    \captionof{figure}{Annotation guidelines for our \textsc{Newsela-Turk} corpus along with  example questions.}
    \label{fig:annotation}
\end{minipage}

\clearpage

\section{System Outputs}
\label{app:sys_out}

\begin{table*}[h!]
\centering
\small
\begin{tabular}{l|p{12.2cm}}
  \hline
  &\textbf{System Outputs} \\
  \hline

Complex   &   \textit{This year, the FAA has approved dozens of permits for agricultural drone businesses.} \\
Simple   &   \textit{This year, it approved dozens of permits for agricultural drone businesses.} \\
\hline
Hybrid-NG   &   \textit{this year, the {\color{blue}\textbf{government}} has approved dozens of drone permits for agricultural businesses.} \\
LSTM   &   \textit{this year, the faa has approved dozens of permits for agricultural drone businesses.} \\
Transformer$_{bert}$   &   \textit{this year, the faa has approved dozens of permits for agricultural businesses.} \\
EditNTS   &   \textit{this year, the {\color{blue}\textbf{government}} has approved dozens of permits for drone businesses {\color{red}\textbf{for no permission}}.} \\
\hline
Our Model ($cp$ = 0.6)   &   \textit{this year, the faa {\color{blue}\textbf{has allowed many businesses to use drones}}.} \\
Our Model ($cp$ = 0.7, 0.8)   &   \textit{this year, the faa has approved dozens of permits for {\color{blue}\textbf{drones}}.} \\
\hline
 \hline
Complex   &   \textit{The room echoed with the sounds of song, the beat of drums, the voices of young men.} \\
Simple   &   \textit{As she spoke, the building echoed with music and the beat of drums.} \\
\hline
Hybrid-NG   &   \textit{echoed the room.} \\
LSTM   &   \textit{the room echoed with the sounds of song, the voices of young men.} \\
Transformer$_{bert}$   &   \textit{the room echoed with the sound of song, the beat of drums, the voices of young men.} \\
EditNTS   &   \textit{the room echoed with the sounds of song, the beat of drums, the voices of young men {\color{red}\textbf{who are hungry and legs}}.} \\
\hline
Our Model ($cp$ = 0.6)   &   \textit{{\color{blue}\textbf{the sound of the room was full of sounds}} of young men and the voices of {\color{red}\textbf{cellos}}.} \\
Our Model ($cp$ = 0.7)   &   \textit{{\color{blue}\textbf{the sound of the room sounded like a lot of music,}} and the voices of young men.} \\
Our Model ($cp$ = 0.8)   &   \textit{{\color{blue}\textbf{the sound of the room sounded}} like a song, the beat of drums, and the voices of young men.} \\
\hline
\hline
\end{tabular}
\caption{Examples of system outputs by our paraphrase generation model and other baselines. Our model generates paraphrase-focused simplifications by skipping the splitting and deletion steps and running only the neural paraphrase generation component. ({\color{red} \textbf{red}} marks the errors; {\color{blue}\textbf{blue}} marks good paraphrases). $cp$ is a soft constraint that denotes the extent of paraphrasing in terms of number of words that can be copied from the input.}
\label{table:sys_out_para}
\end{table*}

\section{Additional Evaluation on Newsela}
\label{app:paraphrase_eval}

\begin{table*}[h]
\small
\centering
\begin{tabular}{l|cccc|cc|ccc|ccc}
\hline
 \textbf{Models} & \textbf{SARI} & \textbf{add} & \textbf{keep}  & \textbf{del} & \textbf{FK} & \textbf{SLen} & \textbf{OLen} & \textbf{CR} & \textbf{\%split} & \textbf{s-BL} &  \textbf{\%new} & \textbf{\%eq}  \\ \hline 
 Complex (input) & 20.6 & 0.0 & 61.7 & 0.0 & 9.2 & 16.9 & 17.0 & 1.0 & 0.0 & 100.0 & 0.0 & 100.0  \\
 Simple (reference)& 94.6 & 93.6 & 91.4 & 98.8 & 8.7 & 17.9 & 17.9 & 1.06 & 0.0 & 48.0 & 29.7 & 0.0 \\
\hline
Hybrid-NG &  35.0 & 2.3 & 52.7 & 50.1 & 7.8 & 16.1 & \textbf{17.0} & \textbf{1.0} & 5.6 & 61.7 & 9.3 & 9.1 \\
Transformer$_{bert}$ &  35.3 & 3.4 & 52.9 & 49.6 & 7.0 & 13.5 & 15.2 & 0.91 & 10.4 & 60.2 & 14.4 & 15.7 \\
EditNTS & 37.7 & 2.0 & 56.4 & 54.5 & 7.6 & 14.2 & 15.5 & 0.93 & 8.7 & 69.0 & 7.1 & 3.5 \\ 
\hline
Our Model & 
37.9 & \textbf{4.4} & 51.3 & 58.0 & 6.7 & 13.6 & 15.3 & 0.92 & 9.7 & \textbf{49.2} & 19.2 & \textbf{0.9} \\
Our Model (no split; $cp$ = 0.6) & 
38.3 & 3.9 & 53.8 & \textbf{57.3} & 7.9 & 16.1 & 16.7 & \textbf{1.0} & \textbf{0.0} & 53.4 & \textbf{20.8} & 3.6 \\
Our Model (no split; $cp$ = 0.7) & 
\textbf{39.1} & 3.7 & 58.5 & 55.2 & 8.3 & 16.2 & 16.8 & \textbf{1.0} & \textbf{0.0} & 67.6 & 12.4 & 11.0  \\
Our Model (no split; $cp$ = 0.8) & 
38.0 & 3.3 & \textbf{60.3} & 50.4 & \textbf{8.5} & \textbf{16.4} & 16.9 & 
\textbf{1.0} & \textbf{0.0} & 76.5 & 8.2 & 20.3 \\
\hline
\end{tabular}
\caption{Automatic evaluation results on a subset of Newsela test set that focuses on paraphrasing (8371 complex-simple sentence with compression ratio $>$ 0.9 and no splits). We control the extent of paraphrasing of our models by specifying the percentage of words to be copied (\textit{cp}) from the input as a soft constraint.}
\label{table:paraphrase_results_auto}
\end{table*}

\clearpage

\section{Human Evaluation Interface}
\label{appendix:human_eval}

\vspace{15pt}
\noindent\begin{minipage}{\textwidth}
   \small
    \centering
    \includegraphics[width=\textwidth]{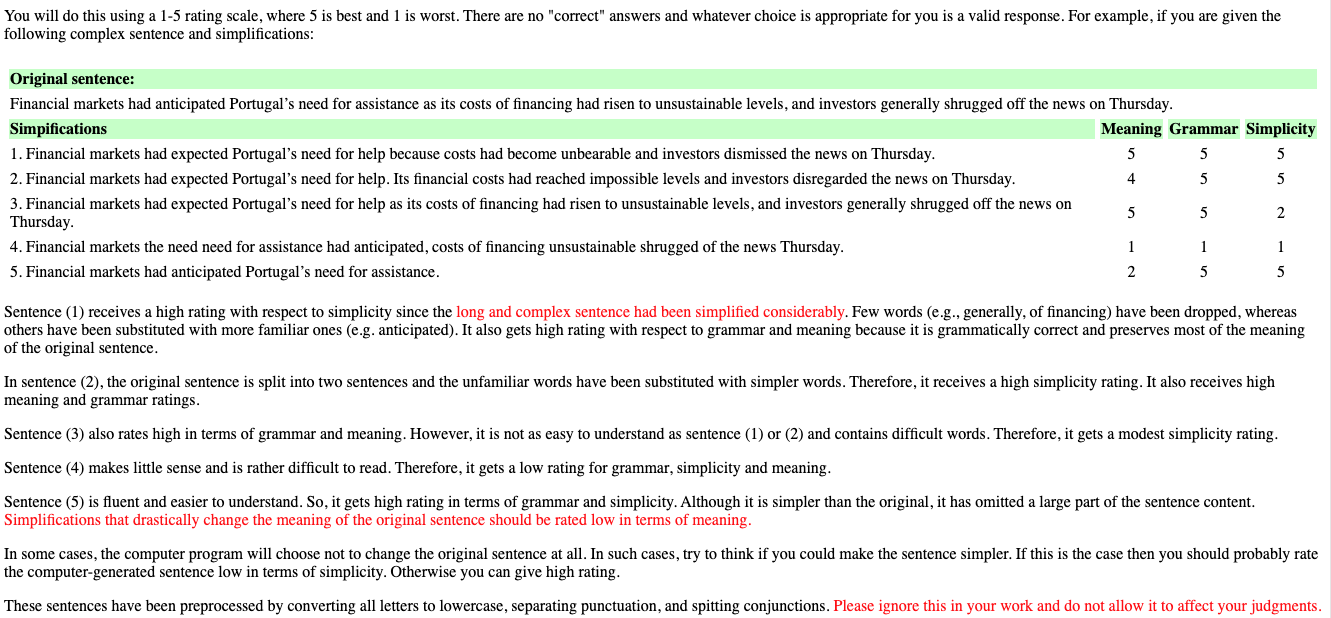}
    \captionof{figure}{Guidelines provided to the Amazon Mechanical Turk workers for evaluating simplified sentences. Our interface is based on the one proposed by \citeauthor{kriz-etal-2019-complexity} \shortcite{kriz-etal-2019-complexity}.}
    \label{fig:evaluation_interface}

\vspace{25pt}

   \small
    \centering
    \includegraphics[width=0.8\textwidth]{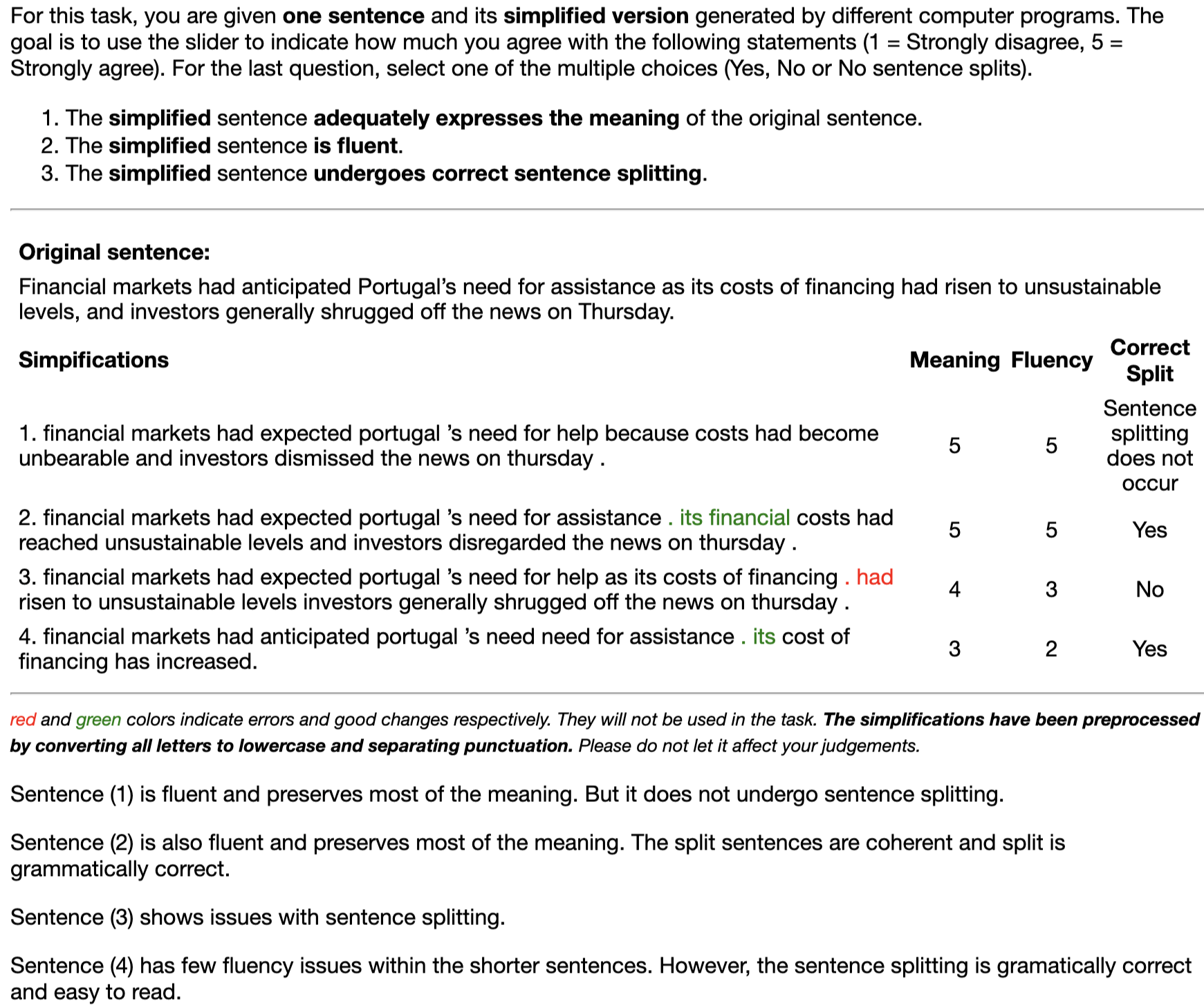}
    \captionof{figure}{Guidelines provided to the Amazon Mechanical Turk workers for evaluating simplified sentences specifically for sentence splitting.}
    \label{fig:split_interface}
\end{minipage}

\clearpage

\section{Evaluation on Wikipedia}
\label{app:Wikipedia}

\begin{table*}[ht]
\small
\centering
\captionsetup{width=\linewidth}
\begin{tabular}{l|cccc|cc|ccc|ccc}
\hline
 \textbf{Models} & \textbf{SARI} & \textbf{add} & \textbf{keep}  & \textbf{del} & \textbf{FK} & \textbf{SLen} & \textbf{OLen} & \textbf{CR} & \textbf{\%split} & \textbf{s-BL} &  \textbf{\%new} & \textbf{\%eq}   \\ \hline 
 Complex (input) & 25.9 & 0.0 & 77.8 & 0.0 & 13.4 & 22.4 & 22.6 & 1.0 & 0.0 & 100.0 & 0.0 & 100.0 \\
 Simple (reference)& 42.0 & 20.6 & 59.9 & 45.5 & 10.9 & 19.1 & 19.3 & 0.88 & 1.1 & 55.2 & 15.3 & 7.8  \\
\hline
Hybrid-NG & 25.4 & 0.1 & 42.7 & 33.5 & 9.0 & 13.3 & 13.4 & 0.6 & \textbf{0.8} & 38.2 & 1.4 & 3.1 \\
LSTM & 32.6 & 2.1 & 59.8 & 36.0 & 10.0 & 17.8 & 17.8 & 0.84 & \textbf{0.8} & 60.0 & 10.7 & 15.0 \\
Transformer$_{bert}$ & 35.1 & 4.3 & \textbf{61.8} & 39.2 & 10.4 & 16.7 & 18.8 & 0.85 & 10.9 & 62.1 & 11.1 & 11.1 \\
EditNTS & 36.1 & 2.5 & 67.4 & 38.5 & 11.7 & 20.9 & 22.4 & 1.02 & 6.4 & 63.5 & \textbf{13.5} & 0.0 \\
\hline
Our Model & 
35.9 & 4.7 & 63.6 & 39.6 & 9.2 & 14.7 & \textbf{19.8} & 0.9 & 33.7 & 63.2 & 12.9 & 9.2 \\
Our Model (no split; $cp$ = 0.6) & 
36.5 & \textbf{4.9} & 63.2 & \textbf{41.4} & \textbf{10.8} & \textbf{18.6} & 19.9 & \textbf{0.89} & 6.7 & 61.9 & 12.4 & 3.9  \\
Our Model (no split; $cp$ = 0.7) & 
\textbf{37.5} & 4.3 & 68.8 & 39.4 & 11.2 & 19.1 & 20.9 & 0.94 & 8.9 & 72.6 & 8.6 & 12.3 \\
Our Model (no split; $cp$ = 0.8) & 
37.0 & 3.8 & 72.0 & 35.3 & 11.7 & 19.8 & 21.7 & 0.97 & 8.4 & 80.4 & 6.6 & 24.5\\
\hline
\end{tabular}
\caption{Automatic evaluation results on \textsc{Turk} dataset \cite{Xu-EtAl:2015:TACL} that focuses on lexical  paraphrasing.}
\label{table:results_turk}
\end{table*}

\begin{table*}[ht]
\small
\centering
\begin{tabular}{p{3.8cm}|cccc|cc|ccc|ccc}
\hline
 \textbf{Models} & \textbf{SARI} & \textbf{add} & \textbf{keep}  & \textbf{del} & \textbf{FK} & \textbf{SLen} & \textbf{OLen} & \textbf{CR} & \textbf{\%split} & \textbf{s-BL} &  \textbf{\%new} & \textbf{\%eq}  \\ \hline 
Complex (input) &  20.5 & 0.0 & 61.5 & 0.0 & 13.4 & 22.4 & 22.6 & 1.0 & 0.8 & 100.0 & 0.0 & 100.0 \\
Simple (reference) &  46.3 & 20.0 & 51.0 & 67.9 & 9.1 & 14.8 & 18.9 & 0.87 & 24.2 & 46.2 & 20.5 & 0.6  \\
\hline
Hybrid-NG & 29.8 & 0.1 & 37.0 & 52.2 & 9.0 & 13.3 & 13.4 & 0.6 & 0.8 & \textbf{38.2} & 1.4 & 3.1\\
LSTM & 36.1 & 2.4 & \textbf{51.8} & 54.2 & 10.0 & 17.8 & 17.8 & 0.84 & 0.8 & 59.9 & 10.8 & 14.8 \\
Transformer$_{bert}$ & 38.7 & 5.0 & 53.5 & 57.7 & 10.4 & 16.7 & \textbf{18.8} & \textbf{0.85} & 10.9 & 62.1 & 11.2 & 11.1  \\
EditNTS & 37.8 & 2.7 & 56.0 & 54.9 & 11.7 & 20.9 & 22.4 & 1.02 & 6.4 & 63.6 & 13.4 & \textbf{0.0} \\
\hline
Our Model & 
 \textbf{39.7} & \textbf{5.3} & 55.1 & \textbf{58.8} & \textbf{9.2} & \textbf{14.7} & 19.8 & 0.9 & \textbf{33.7} & 63.1 & \textbf{14.0} & 8.9 \\
\hline
\end{tabular}
\caption{Automatic evaluation results on \textsc{asset} \cite{fern2020asset} dataset that contains all the three simplification operations.}
\label{table:results_asset}
\end{table*}

\begin{multicols}{2}
We use the complex-simple sentence pairs from {\sc Wiki-auto} \cite{ACL-2020-chao}, which contains 138,095 article pairs and 604k non-identical aligned and partially-aligned sentence pairs. To capture sentence splitting, we join the sentences in the simple article mapped to the same sentence in the complex article. Similar to Newsela, we remove the sentence pairs with high ($>$0.9) and low ($<$0.1) BLEU \cite{Papineni:2002:BMA:1073083.1073135} scores. For validation and testing purposes, we use the following two corpora: (i) \textsc{Turk} corpus \cite{Xu-EtAl:2015:TACL} for lexical paraphrasing and (ii) \textsc{Asset} corpus \cite{fern2020asset} for multiple rewrite operations. While the former corpus has 8 human-written references for 2000 validation and 359 test sentences, the latter corpus provides 10 references for the same sentences. We remove the validation and test sentences from the training corpus. Tables \ref{table:results_turk} and \ref{table:results_asset} show the results on \textsc{Turk} and \textsc{Asset} respectively.
\end{multicols}